\documentclass[letterpaper, 10 pt, conference]{ieeeconf}  % Comment this line out if you need a4paper

\IEEEoverridecommandlockouts                              % This command is only needed if 
\usepackage{amsmath} % assumes amsmath package installed
\usepackage{amssymb}  % assumes amsmath package installed
\usepackage{graphicx}
\usepackage{subcaption}
\usepackage{booktabs}
\usepackage{graphbox}
\usepackage{tikz}
\usetikzlibrary{shapes,arrows,calc,positioning}

\usepackage{todonotes}

\usepackage{soul}

\newcommand{\bez}{B\'ezier~}

\title{\LARGE \bf
Continuous World Coverage Path Planning for Fixed-Wing UAVs\\ using Deep Reinforcement Learning
}

\author{Mirco Theile$^{1*}$, Andres R. Zapata Rodriguez$^{1*}$, Marco Caccamo$^{1}$, Alberto L. Sangiovanni-Vincentelli$^{2}$% <-this % stops a space
\thanks{*These authors contributed equally to this work.}% <-this % stops a space
\thanks{$^{1}$Mirco Theile, Andres R. Zapata Rodriguez, and Marco Caccamo are with TUM School of Engineering and Design, Technical University of Munich, Germany
        {\tt\small \{mirco.theile, andres.zapata-rogdriguez, mcaccamo\}@tum.de}}%
\thanks{$^{2}$Alberto L. Sangiovanni-Vincentelli is with Dept. of Electrical Engineering and Computer Sciences, University of California, Berkeley, USA,
        {\tt\small alberto@berkeley.edu}}%
\thanks{Marco Caccamo was supported by an Alexander von Humboldt Professorship endowed by the German Federal Ministry of Education and Research.}% <-this % stops a space
}

\begin{document}

\maketitle
\thispagestyle{empty}
\pagestyle{empty}

%%%%%%%%%%%%%%%%%%%%%%%%%%%%%%%%%%%%%%%%%%%%%%%%%%%%%%%%%%%%%%%%%%%%%%%%%%%%%%%%
\begin{abstract}
Unmanned Aerial Vehicle (UAV) Coverage Path Planning (CPP) is critical for applications such as precision agriculture and search and rescue. While traditional methods rely on discrete grid-based representations, real-world UAV operations require power-efficient continuous motion planning. We formulate the UAV CPP problem in a continuous environment, minimizing power consumption while ensuring complete coverage. Our approach models the environment with variable-size axis-aligned rectangles and UAV motion with curvature-constrained B\'ezier curves. We train a reinforcement learning agent using an action-mapping-based Soft Actor-Critic (AM-SAC) algorithm employing a self-adaptive curriculum. Experiments on both procedurally generated and hand-crafted scenarios demonstrate the effectiveness of our method in learning energy-efficient coverage strategies.
\end{abstract}

\section{Introduction}

Unmanned Aerial Vehicle (UAV) Coverage Path Planning (CPP) is a challenging problem with numerous real-world applications. It is critical in precision agriculture, search and rescue missions, environmental monitoring, infrastructure inspection, surveying, and mapping. Fixed-wing UAVs, in particular, are of interest due to their efficient power consumption, making them well-suited for long-distance missions.

CPP is an NP-hard problem~\cite{arkin2000approximation}. In contrast to the well-known Traveling Salesperson Problem (TSP), which focuses on visiting discrete locations, CPP requires areas to be covered by a sweeping area determined by the robot's location. Consequently, every point in the target area can be covered from any location within an area the size of the sweeping area, making the problem especially difficult to formulate and solve in continuous environments. This challenge is further exacerbated for UAV CPP since the sweeping area is usually determined by the Field of View (FoV) of a camera mounted underneath the UAV, producing a significantly larger sweeping area than for most ground robots.

A prominent survey has analyzed the CPP literature~\cite{choset_coverage_2001}. Its recent counterparts, extensions \cite{galceran_survey_2013,cabreira_survey_2019} and systematic reviews \cite{kumar_region_2023}  presented taxonomies, classifying CPP techniques, and established its importance in various domains. Traditional CPP approaches frequently assume a discretized world representation, where the environment is modeled as a grid or a graph. This simplification enables the application of classical search algorithms and optimization techniques. However, this discretization introduces limitations in accuracy and efficiency, since the real world is inherently continuous.

\begin{figure}
\centering
\begin{minipage}{0.49\columnwidth}
\includegraphics[width=\textwidth]{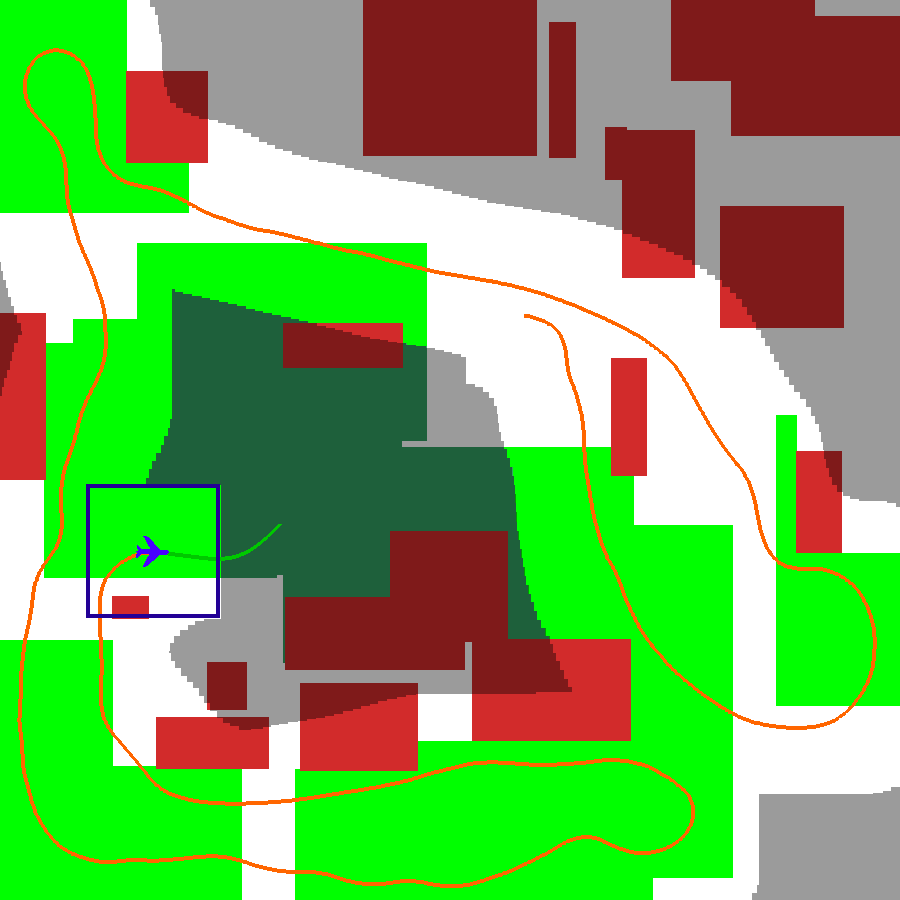}
\end{minipage}%
\hfill%
\begin{minipage}{0.50\columnwidth}
\small
\renewcommand{\arraystretch}{1.25}
\setlength{\tabcolsep}{4pt}
\begin{tabular*}{\textwidth}{cl}
\toprule[1.5pt]
 & Description\\
\midrule
\includegraphics[align=c,height=.3cm]{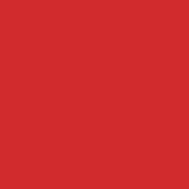} & No-fly zones (NFZs)\\
\includegraphics[align=c,height=.3cm]{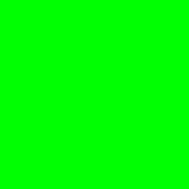} & Target zones \\
\includegraphics[align=c,height=.3cm]{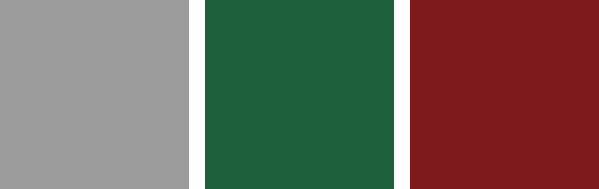} & Not covered areas\\
\includegraphics[align=c,height=.3cm]{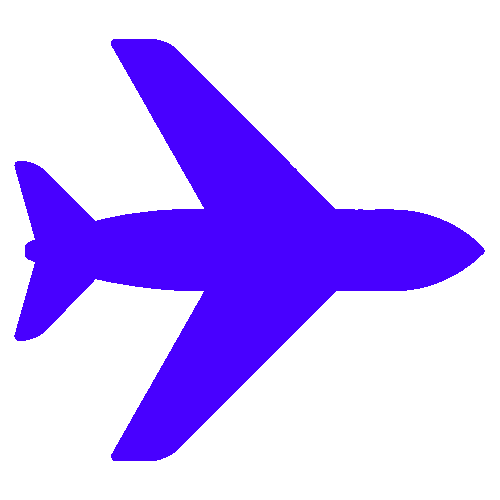} & Agent\\
\includegraphics[align=c,height=.3cm]{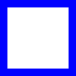} & Field of view (FoV)\\
\includegraphics[align=c,width=.5cm]{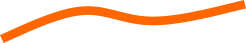} & Trajectory\\
\includegraphics[align=c,width=.5cm]{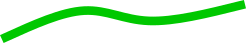} & Action\\
\bottomrule[1.5pt]
\end{tabular*}
\end{minipage}
\caption{Example state of a fixed-wing UAV in a continuous world CPP problem on the left, showing the covered area, trajectory, and field of view, with a legend on the right.}
\label{fig:cpp}
\end{figure}

The continuous real world does not conform to a predefined grid structure, and UAVs do not occupy discrete cells, but move through smooth, continuous trajectories. Coverage optimization in continuous environments involves generating smooth paths rather than visiting a predefined set of discrete waypoints. This shift in representation introduces new challenges in both modeling and optimization. A related paper \cite{coombes_fixed_2018} tackled continuous CPP by decomposing complex target zones into convex polygons and employing boustrophedon heuristics with optimized turn-radius calculations for energy-efficient coverage. \cite{bircher_receding_2018} proposed a structural inspection planner that utilizes an online Rapidly-exploring-Random-Tree (RRT)-based receding-horizon approach, which does not explicitly address energy minimization or nonholonomic constraints, targeting quadcopter platforms. Despite these advances, there is still a significant gap in the integration of dynamic constraints, energy efficiency, and real-time adaptability into a unified CPP framework.

Recent advances have drawn attention to learning-based methods for solving path planning problems. Deep Reinforcement Learning (DRL), in particular, has gained traction due to its ability to learn solutions without explicit supervision, relying on trial-and-error interactions. Recent papers focused on applying DRL to CPP in discretized grid-world environments~\cite{theile2023learning,carvalho_deep_2025}.

However, RL faces significant challenges, such as ensuring safety and high sample complexity. Furthermore, long-horizon planning problems remain particularly difficult for RL-based approaches~\cite{li2022challenges,lehnert2018value}. The most recent DRL publications for UAVs focus primarily on quadcopters. However, there is a growing need for solutions tailored to fixed-wing platforms, whose advantages in large-scale outdoor missions have sparked considerable interest \cite{richter2024review}. To the best of our knowledge, most existing methods still rely on discrete-action Q-learning \cite{richter2024review,yan2020towards}, typically addressing point-to-point navigation or obstacle avoidance. These methods do not naturally extend to long-horizon coverage tasks, leaving largely unexplored integrated coverage objectives, energy efficiency, and nonholonomic dynamics.

In this paper, we formulate the continuous-world coverage optimization problem with obstacle avoidance and nonholonomic dynamical constraints to minimize power consumption while achieving full coverage of target zones. We model the environment using variable-size sets of axis-aligned rectangles to approximate real-world conditions. Additionally, we represent the fixed-wing UAV's motion using smooth, curvature-constrained \bez curves, which, under ideal conditions, can be perfectly followed by an underlying autopilot, such as uavAP~\cite{theile2020uavap}. Fig. ~\ref{fig:cpp} illustrates the problem.

To solve this problem, we introduce a set-based environment representation, employing a custom attention-based neural network to process spatial information. Our action space is designed to generate \bez curves, ensuring continuity in position, velocity, and curvature. Furthermore, we define a feasibility model for the environment and train an action-mapping-based RL agent (AM-SAC)~\cite{theile2024action}. We implement a self-adaptive curriculum to facilitate efficient learning, allowing the agent to progressively advance from simpler to more complex scenarios based on performance metrics.

We validate our approach through extensive experiments, demonstrating its effectiveness by analyzing training curves and showcasing the adaptive difficulty adjustments. Our evaluation includes testing multiple trained agents on randomly generated scenarios and visualizing selected trajectories on procedurally generated and hand-crafted maps.

The main contributions of this work are as follows:
\begin{itemize}
    \item Formulation of the continuous UAV complete CPP problem with an objective of power minimization.
    \item Utilization of feasibility models and curriculum learning to train a DRL agent capable of solving procedurally generated scenarios.
    \item Evaluation of the effectiveness of our curriculum-based training and the final agent's performance on both generated and hand-crafted scenarios.
\end{itemize}

\section{Preliminaries}
\subsection{State-wise Constrained Markov Decision Process}
A State-wise Constrained Markov Decision Process (SCMDP) \cite{zhao2023state} is defined by the tuple $ (\mathcal{S}, \mathcal{A}, \mathrm{R}, \{\mathrm{C}_i\}_{\forall i}, \mathrm{P}, \gamma, \mathcal{S}_0, \mu) $, where $ \mathcal{S} $ and $ \mathcal{A} $ denote the state and action spaces, respectively. The reward function $ \mathrm{R}:\mathcal{S}\times \mathcal{A}\to \mathbb{R} $ specifies the immediate reward received for executing an action in a given state. The system's stochastic evolution is governed by the transition function $ \mathrm{P}:\mathcal{S}\times \mathcal{A}\to\mathcal{P}(\mathcal{S}) $, where $ \mathcal{P}(\mathcal{S}) $ represents a probability distribution over the state space. The discount factor $ \gamma \in [0,1] $ balances the influence of immediate versus future rewards. Additionally, $ \mathcal{S}_0 $ defines the set of initial states, and $ \mu = \mathcal{P}(\mathcal{S}_0) $ represents the initial state distribution. Given a stochastic policy $ \pi:\mathcal{S}\to\mathcal{P}(\mathcal{A}) $, the expected discounted cumulative reward is expressed as  
\begin{equation} \label{eq:rl_objective}
\resizebox{0.9\linewidth}{!}{$
    \mathrm{J}(\pi) = \mathbb{E} \left[ \sum_{t=0}^{\infty} \gamma^t \mathrm{R}(s_t, a_t) \,\Big|\, 
    s_0\sim \mu, a_t\sim\pi, s_{t+1} \sim \mathrm{P} \right].$}
\end{equation}

Unlike a standard Markov Decision Process (MDP)~\cite{sutton1998reinforcement}, an SCMDP incorporates a set of cost functions $ \{\mathrm{C}_i\}_{\forall i} $, where each function $ \mathrm{C}_i: \mathcal{S}\times\mathcal{A}\times\mathcal{S}\to \mathbb{R} $ assigns a cost to every transition. Considering all possible trajectories $ \tau^\pi(s) $ generated by policy $ \pi $ from an initial state $ s $, the SCMDP optimization problem is formulated as  
\begin{align} \label{eq:const_obj}
    \pi^* = & \operatorname{arg}\max_{\pi\in\Pi} \mathrm{J}(\pi)  \\
    \text{s.t.} \quad & \mathrm{C}_i(s_t, a_t, s_{t+1}) \leq w_i, \quad \forall i,  \nonumber \\
    & (s_t, a_t, s_{t+1})\sim \tau^\pi(s_0), \quad \forall s_0\in\mathcal{S}_0. \nonumber
\end{align} 
This formulation ensures that for every transition $ (s_t, a_t, s_{t+1}) $ encountered in any possible trajectory of $ \pi $, the associated cost function $ \mathrm{C}_i $ remains within the specified bound $ w_i $, across all possible initial states in $ \mathcal{S}_0 $.

\subsection{Reinforcement Learning}
Finding a policy $\pi$ that maximizes the cumulative discounted reward of an MDP is the goal of Reinforcement Learning.
The Soft Actor-Critic (SAC) algorithm is widely used for continuous control. SAC is an off-policy actor-critic algorithm with maximum entropy optimization~\cite{haarnoja2018soft}.
SAC maximizes a reward signal while encouraging exploration through an entropy regularization term. The SAC objective is given by: 
\begin{equation} \
\mathrm{J}(\pi) = \mathbb{E}_{(s_t,a_t)\sim \pi} \left[Q(s_t, a_t) + \alpha\, \mathcal{H}\left(\pi(\cdot | s_t)\right)\right],
\end{equation} 
where $Q(s_t, a_t)$ estimates the value of performing action $a_t$ at state $s_t$, $\mathcal{H}(\pi(\cdot| s_t))$ is the entropy of the policy, and $\alpha>0$ controls the trade-off between maximizing return and encouraging exploration.

\subsection{Action Mapping}

Action Mapping (AM)~\cite{theile2024action} is a model-based technique to address SCMDPs by decoupling constraint satisfaction from objective-driven action selection. Using a feasibility model $g: \mathcal{S} \times \mathcal{A \to \mathbb{B}}$ that distinguishes between feasible and infeasible actions, a \textit{feasibility policy} $\pi_f$ is pre-trained to generate all feasible actions for given states~\cite{theile2024learning}. After the pre-training, an \textit{objective policy} $\pi_o$ is trained with standard RL techniques---such as SAC, yielding AM-SAC---to choose the optimal action among the feasible ones. 

Ideally, if a perfect feasibility policy is pre-trained, the SCMDP is converted into an unconstrained MDP. However, since the feasibility policy is learned and relies on an approximate feasibility model, constraint satisfaction is not guaranteed. Nevertheless, AM can still drastically reduce the number of infeasible actions, which can significantly improve the training of the objective policy.

\subsection{B\'ezier Curves}
Since fixed-wing UAVs must generate smooth, feasible trajectories, we use \bez curves to model their motion while satisfying curvature constraints.
A B\'ezier curve \cite{bezier1977essai} of degree $n$ is defined as:
\begin{equation}
\mathbf{p}(u) = \sum_{i=0}^{n} B_i^n (u) \mathbf{b}_i, \quad u \in [0,1]
\end{equation}
where $\mathbf{b}_i$ are the control points and $B_i^n(u)$ are the Bernstein polynomials given by:
\begin{equation}
B_i^n (u) = \binom{n}{i} u^i (1 - u)^{n-i}. 
\end{equation}
The first and second derivatives of the B\'ezier curve are
\begin{equation}
\mathbf{p}'(u) = \sum_{i=0}^{n-1} B_i^{n-1} (u) \cdot n(\mathbf{b}_{i+1} - \mathbf{b}_i),
\end{equation}
\begin{equation}
\mathbf{p}''(u) = \sum_{i=0}^{n-2} B_i^{n-2} (u) \cdot n (n-1) (\mathbf{b}_{i+2} - 2\mathbf{b}_{i+1} + \mathbf{b}_i),
\end{equation}
which can be used to derive the direction and curvature along the spline.
\section{Problem Formulation}

In this paper, we consider a fixed-wing UAV equipped with a gimbal-mounted camera that is tasked to cover designated target zones while avoiding no-fly zones. Fig.~\ref{fig:cpp} illustrates the problem. The following specifies the environment-UAV interaction and lays out the optimization objective.

\subsection{Environment}
We consider a two-dimensional rectangular environment containing only axis-aligned rectangular elements, which we define through the set $\mathcal{R}$. The environment contains two types of elements: no-fly zones (NFZs) and target zones (TZs).

The UAV is not allowed to enter a set of $N$ rectangular NFZs, which are static over an episode.
\begin{equation}
    \mathcal{N} = \{\mathbf{N}_i | \mathbf{N}_i \in \mathcal{R}, \forall i \in [1, N]\}
\end{equation}

Additionally, the environment contains $C(t)$ rectangular TZs
\begin{equation}
    \mathcal{C}(t) = \{\mathbf{C}_j | \mathbf{C}_j \in \mathcal{R}, \forall j \in [1, C(t)]\}
\end{equation}
that the UAV needs to cover with its FoV. Once covered, the TZs are removed from the set. If partially covered, target rectangles are split to represent the remaining area. Therefore, the area and the number of TZs are time-dependent. 

\subsection{Fixed-Wing UAV}
The UAV flies with constant speed and has a maximum roll angle, limiting the maximum turning curvature. This paper assumes that the UAV is controlled by a perfect trajectory-tracking controller that can follow polynomial spline path sections that do not violate the curvature constraint. A possible autopilot capable of following spline path sections is the uavAP autopilot \cite{theile2020uavap}.

With this assumption, we define the motion of the UAV as a sequence of B\'ezier curves $\mathbf{p}_k(u)$, switching every $T_b$ seconds, having smooth transitions, i.e., continuity of position, velocity, and curvature. As we assume the UAV moves with constant Cartesian speed $v_\text{const}$, the B\'ezier parameter $u(t)$ needs to satisfy
\begin{equation}\label{eq:dudt}
\frac{du}{dt} = \frac{v_\text{const}}{|\mathbf{p}_k'(u)|}, \qquad u(kT_b) = 0.
\end{equation}
At every time $t = kT_b$, the next B\'ezier curve starts, and $u$ resets.
The Cartesian position of the UAV is then given by
\begin{equation}
\mathbf{x}(t) = \mathbf{p}_{\lfloor t/T_b \rfloor} ( u(t) ),
\end{equation}
where $k = \lfloor t/T_b \rfloor$ determines the active B\'ezier curve and $u(t)$ evolves dynamically within each segment according to~\eqref{eq:dudt}.
The curvature of the UAV's trajectory at any time $t$ is given by
\begin{equation}
\kappa(t) = \frac{x_k'(u) y_k''(u) - y_k'(u) x_k''(u)}{\left( x_k'(u)^2 + y_k'(u)^2 \right)^{3/2}} \Bigg|_{u = u(t)}
\end{equation}
The required roll angle $\phi(t)$ can be approximated from the curvature along the trajectory using the steady-state turn relationship
\begin{equation}
    \phi(t) = \tan^{-1} \left( \frac{v_\text{const}^2 \kappa(t)}{g} \right),
\end{equation}
which is derived from the required induced lift to perform a turn with curvature $\kappa(t)$.

\subsection{Evolution of the Target Zones}
The UAV is equipped with a gimbal-mounted camera, which is assumed to have a square, axis-aligned field of view (FoV). The size of the FoV square is determined by the altitude of the UAV $h$ and the camera's angle of view $\alpha$. As such, the side length of the square is computed as 
\begin{equation}
    s = 2h\tan\left(\frac{\alpha}{2}\right).
\end{equation}
The UAV's constant altitude assumption allows us to model a UAV-centered FoV with constant area.
Therefore, the FoV can be expressed as a function of the UAV position as $V:\mathbb{R}^2\to\mathcal{R}$.

The camera captures a frame every $T_f$ seconds. TZs within the FoV are removed or cropped according to the overlap. Therefore, at frame $l+1$, the TZs can be described as
\begin{equation}
    \mathcal{C}_{l+1} = \mathcal{C}_l \setminus V(\mathbf{x}((l+1)T_f)).
\end{equation}
While the number of TZs $C(t)$ can increase to represent partially covered rectangles, the total area is monotonically decreasing.

\subsection{Power Modeling}
The employed power model is derived from \cite{dantsker2018high}, where the instantaneous power consumption at the battery is given by:
\begin{equation}
    P(t) = \frac{K_i}{\eta_m\eta_p}\frac{\cos^2\gamma}{v\cos^2\phi(t)} + \frac{K_p}{\eta_m\eta_p}v^3 + \frac{m}{\eta_m\eta_p}(g\sin\gamma + a)v,
\end{equation}
which divides power into lift-induced, parasitic, and dynamic components with motor and propeller efficiency factors $\eta_m$ and $\eta_p$. With the given assumptions of constant altitude and speed, i.e., no climb angle $\gamma$ nor forward acceleration $a$, the model can be simplified to 
\begin{equation}
    P(t) = A\frac{1}{v\cos^2\phi(t)} + Bv^3.
\end{equation}
Parameters $A$ and $B$ can be regressed from data as done in \cite{theile2018uavee}. Given a constant speed $v_\text{const}$, the power consumption only varies based on the roll angle $\phi(t)$ and, thus, depending on the curvature of the spline segments. 

\subsection{Optimization Objective}
The problem addressed in this paper is to minimize the power consumption of a complete coverage path planning problem with a roll-angle-constrained fixed-wing UAV. Mathematically, it can be described as
\begin{align}
    \min_{\mathbf{p}_k, \forall k \in [0, \lfloor T/T_b\rfloor]} &\int_0^T P(t)dt \label{eq:objective}\\
    \text{s.t.}\quad & C_{\lfloor T/T_f\rfloor} = \{\}  \label{eq:const_coverage}\\
    & \mathbf{x}(t) \notin \mathcal{N}, \forall t\in [0, T] \label{eq:const_nfz}\\
    & \phi(t) \in [-\phi_\text{max}, \phi_\text{max}], \forall t\in [0, T] \label{eq:const_curve}\\
    & \mathbf{x}(t), \mathbf{v}(t), \kappa(t)\in C^0([0, T])\label{eq:const_smooth} .
\end{align}
The objective is to select the B\'ezier curves $\mathbf{p}_k$ that minimize power consumption while covering all TZs until the last path section, as described in \eqref{eq:const_coverage}. The constraint~\eqref{eq:const_nfz} specifies that the UAV is not allowed to enter any of the NFZs at any time, and the constraint~\eqref{eq:const_curve} limits the roll angle and consequently the curvature of the curves to make them achievable by the underlying autopilot. The final constraint~\eqref{eq:const_smooth} specifies that the position, velocity, and curvature are $C^0$ continuous, which enforces smoothness when transitioning between B\'ezier curves. 
\section{Methodology}

We solve the continuous-world CPP problem using DRL. The following transforms the optimization objective into an SCMDP, presents a curriculum enabling the DRL agent to learn how to minimize power consumption, and details the neural network used.

\subsection{SCMDP Formulation}

The state of the environment and of the UAV can be described through the size of the environment, lists of NFZ rectangles and target rectangles, the current position, velocity, and curvature of the UAV, and the time since the last camera frame was taken. To represent this state for a deep neural network (DNN)-based agent, we rearrange it by formulating an observation of the state.

\subsubsection{Observation Space}

The state values position, velocity, curvature, environment shape, and time since the last camera frame are normalized and concatenated to a \mbox{$o_\text{scalars}\in\mathbb{R}^8$} observation. 

The NFZs are represented as an unordered set of $N$ rectangles, where each rectangle is described through its top-left and bottom-right coordinates. The two coordinates are presented through their offsets to the top-left and bottom-right coordinates of the map and as an offset relative to the UAV's current position. Additionally, the area of the rectangle is provided, yielding a \mbox{$o_\text{NFZs}\in\mathbb{R}^{N \times 13}$}.

\begin{figure}
    \centering
    \vspace{5pt}
    \resizebox{0.8\columnwidth}{!}{\input{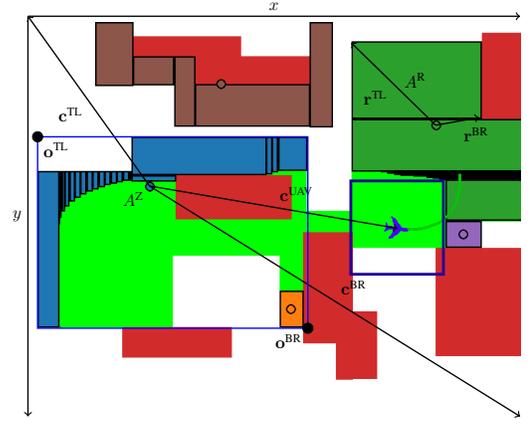}}
    \caption{Illustration of the TZ observation, showing five TZs in different colors and the rectangles that comprise them. Circles in the center indicate the centroids of the zones. Measurements in the green zone show the rectangle descriptors within a zone, while the ones for the blue zone indicate the TZ descriptors.}
    \label{fig:target_obs}
\end{figure}

While the target rectangles could be represented like the NFZs, this representation would be challenging for the DNN as the number of TZ can grow rapidly depending on the UAV's trajectory. 
Fig.~\ref{fig:target_obs} shows how many rectangles are needed to describe the remaining TZ after the UAV passes over some zones. Therefore, we group the target rectangles into connected zones as illustrated by the different colors in Fig.~\ref{fig:target_obs}. Each connected zone is then described through a list of rectangles with its respective zone descriptor. 
The rectangles are described by their top-left and bottom-right coordinates relative to the zone centroid, $\mathbf{r}^\text{TL}$ and $\mathbf{r}^\text{BR}$, and their area $A^\text{R}$. 

The zone descriptor contains the offset of the centroid coordinate relative to the top-left and bottom-right of the environment, $\mathbf{c}^\text{TL}$ and $\mathbf{c}^\text{BR}$ and the centroid's offset to the UAV $\mathbf{c}^\text{UAV}$. Additionally, the top-left and bottom-right corners of the outer rectangle, $\mathbf{o}^\text{TL}$ and $\mathbf{o}^\text{BR}$ relative to the environment's top-left and bottom left coordinate each are observed.
Additionally, the total area of the zone $A^\text{Z}$ is added to the descriptor. Therefore, each of the $Z$ connected TZ is described through $o_\text{TZ}^\text{R} \in \mathbb{R}^{R \times 5}$, with $R$ being the number of rectangles in the connected zone and $o_\text{TZ}^\text{D} \in \mathbb{R}^{15}$ for the zone descriptor. 

\subsubsection{Action Space}

\begin{figure}
    \centering
    \vspace{5pt}
    \includegraphics[width=0.5\linewidth]{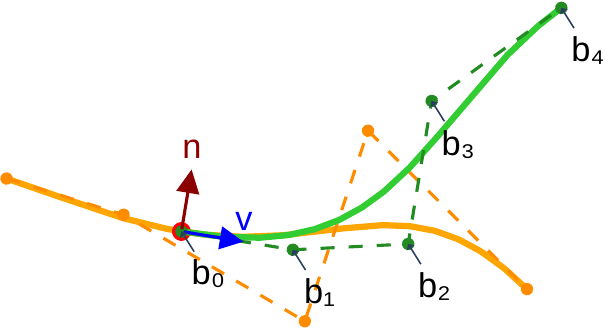}
    \caption{Visualization of the transition between consecutive quartic \bez curves, highlighting the control points for the next spline (green) and the direction and normal vector of the UAV.}
    \label{fig:action_space}
\end{figure}

Every $T_b$ seconds, the agent decides the next \bez curve for the UAV to follow, as visualized in Fig.~\ref{fig:action_space}. The agent provides an action vector $\mathbf{a}\in[-1,1]^{|\mathcal{A}|}$, which is translated into the control points of the \bez curve. To ensure continuity in position, velocity, and curvature, the three initial points $\mathbf{b}_0,\mathbf{b}_1,$ and $\mathbf{b}_2$ of the \bez curve need to be constrained as follows:
\begin{equation}
   \mathbf{b}_0 = \mathbf{x}(t),
\end{equation}
ensures that the next \bez curve starts at the current position. The next point needs to be along the current direction of the UAV to ensure velocity continuity through
\begin{equation}
    \mathbf{b}_1 = \mathbf{b_0} + \frac{a_0 + 1}{2}\lambda\Bar{\mathbf{v}}(t),
\end{equation}
in which $\Bar{\mathbf{v}}(t)$ is the normalized velocity vector, $a_0$ is the first component of the agent's action, and $\lambda$ is an action space scaling factor. The initial curvature at the next spline is only affected by the components of the first three points orthogonal to the UAV's current direction. Since the orthogonal components of the first two points are already defined through position and velocity continuity, the third point's orthogonal component is determined by the current curvature $\kappa(t)$. As such, the third point is given by
\begin{equation}
    \mathbf{b}_2 = 2\mathbf{b}_1 - \mathbf{b}_0 + \frac{n}{n-1}\kappa(t)||\mathbf{b}_1 - \mathbf{b}_0||^2 \mathbf{n}(t) + a_1 \lambda \Bar{\mathbf{v}}(t),
\end{equation}
in which $\mathbf{n}(t)$ is the normalized normal vector to the current direction of the UAV. All remaining points of the \bez curve are then defined as
\begin{equation}
    \mathbf{b}_i = \mathbf{b}_0 + \lambda(a_{2i-4}\Bar{\mathbf{v}}(t) + a_{2i-3}\mathbf{n}(t)), \quad \forall i > 2
\end{equation}
In this work, we utilize quartic \bez curves that require five control points, thus yielding an action space of $\mathcal{A}=[-1,1]^6$.

\subsubsection{Feasibility Model}
The action space definition enforces the continuity constraint in~\eqref{eq:const_smooth}. However, the NFZ and roll angle inequality constraints~\eqref{eq:const_nfz} and~\eqref{eq:const_curve} cannot be easily enforced through specific action space definitions. Since these constraints would be challenging to learn by a standard DRL algorithm, as shown in~\cite{theile2024action}, we utilize action mapping to learn the constraints and then the objective subsequently. To train the feasibility policy $\pi_f$, an approximate feasibility model $g:\mathcal{S}\times\mathcal{A}\to\mathbb{B}$ is needed. The feasibility model assesses the feasibility of an action at a given state. 

In this work, we utilize a similar feasibility model as discussed in~\cite{theile2024action}. Specifically, we create the spline from the state and action as discussed before, generate points on the spline, and assess the following:
\begin{itemize}
    \item Is the local curvature $\kappa$ at any point exceeding the maximum allowed curvature $\kappa_\text{max}$ derived from $\phi_\text{max}$;
    \item Is any of the points outside the map or inside an NFZ;
    \item Is the cumulative distance between the points, i.e., the approximate length of the spline outside a predefined range?
\end{itemize}
If either of these is true, the defined action is infeasible; otherwise, it is feasible. The predefined range in this work is 2.5-3.5 times the distance the UAV covers in $T_b$ seconds. The last check is done so that the spline describes the next $T_b$ seconds of the UAV's motion and represents a feasible continuation of that motion.

\subsubsection{Reward Function}
We chose the reward function aligned with the optimization objective simply as a penalty for power consumption and a single reward when finishing the task. Even though action mapping significantly reduces the constraint violations of the agent, it does not strictly enforce constraint satisfaction. Therefore, when a constraint is violated, we reset the agent to the state before the violating action and give it a penalty equivalent to flying a curve with the highest allowed curvature. If the agent is reset multiple times in a row, the episode is truncated, i.e., reset without setting the termination flag. We chose this method of dealing with constraint violations to still enforce the hard constraints---the agent cannot violate them to achieve the goal---but without giving the agent a way to avoid the long-term power penalty.

The simple reward function has the benefit that maximizing the return also maximizes the objective. Since there needs to be a discount factor for stabilizing training, including a reward for coverage leads to a greedy behavior of the agent. Therefore, we do not yield a coverage reward. However, omitting the coverage reward means that the reward has no instructive value to the agent unless it completes an episode. Therefore, we further designed a curriculum that allows the agent to learn with this simple reward function.

\subsection{Curriculum}

\begin{figure*}
    \centering
    \vspace{5pt}
    \begin{subfigure}{0.5\linewidth}
        \centering
        \includegraphics[height=5cm]{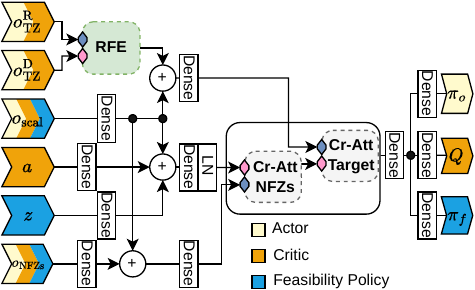}
        \caption{Overall neural network architecture used for all networks, with color-coded input and output differences.}
        \label{fig:nn_overall}
    \end{subfigure}\hfill%
    \begin{subfigure}{0.2\linewidth}
        \centering
        \includegraphics[height=5cm]{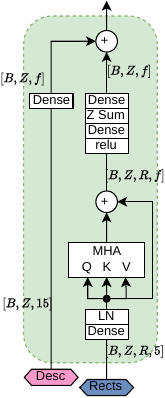}
        \caption{Rectangle Feature Extractor (RFE) module.}
        \label{fig:nn_rfe}
    \end{subfigure}\hspace{20pt}%
    \begin{subfigure}{0.2\linewidth}
        \centering
        \includegraphics[height=5cm]{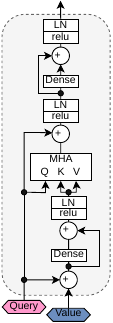}
        \caption{Cross-Attention (Cr-Att) module.}
        \label{fig:nn_cr_att}
    \end{subfigure}
    \caption{Neural network architecture and components.}
    \label{fig:nn_arch}
\end{figure*}
The problem with the simple reward that does not contain a coverage reward is that the agent can only learn when it regularly finishes a task. On a large map, randomly finishing a task is nearly impossible. Therefore, we introduce a curriculum that slowly increases the map's size depending on the agent's performance. 

Formally, we define an episodic task in which episode $k$ has a difficulty parameter $d_k\in(0, 1]$ and is either solved or not solved, indicated through the success variable $s_k\in\{0,1\}$. The learning progress can be defined through a progress variable that evolves according to
\begin{equation}
    p_{k+1} = \tau s_k d_k + (1-\tau) p_k, \qquad p_0 = 0
\end{equation}
as a filtered difficulty-weighted success, with filter parameter $\tau\in(0,1]$. We define a success threshold $s_\text{th}\in(0,1)$ so that the difficulty increases if the success rate for a specific difficulty is above the threshold and decreases if it is below. Consequently, the difficulty for the next episode can be computed as 
\begin{equation}
    d_k = \operatorname{clip}\left(\frac{p_k}{s_\text{th}}, d_\text{min}, 1\right),
\end{equation}
with some minimum difficulty $d_\text{min}\in(0, 1]$. Once the progress exceeds $s_\text{th}$, the difficulty is the maximum difficulty.

In our problem of CPP, the difficulty linearly determines the area of the environment. The effect of this curriculum is illustrated and discussed in Fig.~\ref{fig:curriculum}.

\subsection{Neural Network}

% A key difficulty in designing the DNN for this problem is that the target zones evolve depending on the smooth and continuous motion of the UAV. In particular, large changes in coverage—such as the successful completion of a full zone—can be as important to the agent as small changes resulting from a few leftover rectangles. Moreover, the relative distance of the UAV to these zones does not necessarily correlate with their importance. Near and far target zones can be equally critical for making coverage decisions at a given time. % This can be removed if we need space

Fig.~\ref{fig:nn_arch} illustrates the neural network architecture utilized in this work. The vectorized observations $o_\text{scalars}$ are mixed with the action $a$ (critic only) or the latent $z$ (feasibility policy) to create the query for the following processing. For the connected TZ described through variable size $o_\text{TZ}^\text{R}$ and $o_\text{TZ}^\text{D}$, features are extracted using a rectangle feature extractor (RFE). The NFZ observation $o_\text{NFZs}$ is processed with fully connected (Dense) layers. 

After preprocessing, the vectorized state representation alternatingly attends to the NFZ and the target zone representations, which can be repeated multiple times. All relevant information from the NFZs and TZ can be extracted through these cross-attentions. After the cross-attention, the output is processed through independent Dense layers to create the policy $\pi_o$, critic $Q$, or the feasibility policy $\pi_f$. 

The attention-based architecture is used since the number of NFZs and TZ is variable, and, more crucially, their processing should be permutation invariant. The following elaborates on the modules used.

\subsubsection{Rectangle Feature Extractor (RFE)}
The RFE, visualized in Fig.~\ref{fig:nn_rfe}, exploits the target's spatially coherent separation into zones and generates features that describe each zone individually. First, the rectangles in the zone are processed through self-attention. Then, the features of the rectangles are summed, representing the total feature of the zone. Afterward, they are mixed with the handcrafted zone descriptors.

\subsubsection{Cross-Attention Module (Cr-Att)}
The Cr-Att module in Fig.~\ref{fig:nn_cr_att} allows the DNN to extract relevant information from the variable sets of NFZs or TZs based on a query. After further processing the value input, the query attends to the value features, extracting relevant information. The extracted values are then added to the query residual and further processed through a Dense layer to form the module's output.

\section{Experiments}

\subsection{Simulation Setup}
%=============================
% TABLE: Environment Parameters
%=============================
\begin{table}[t]
    \centering
    \scriptsize
    \begin{tabular*}{0.7\linewidth}{l|c|l}
    \toprule[1.5pt]
    \textbf{Parameter} & \textbf{Value} & \textbf{Description} \\
    \midrule
    \multicolumn{3}{c}{\textbf{World}} \\
    \hline
    $w_\text{max}$ & 2 km & Maximum side length  \\
    $N$ & 20 & No. of NFZs \\
    $Z$ & 10  & No. of initial TZs \\
    \midrule
    \multicolumn{3}{c}{\textbf{UAV Parameters}} \\
    \hline
    $v_\text{const}$ & 20\,m/s & Constant velocity \\
    $\phi_\text{max}$ & 45\textdegree & Maximum roll angle \\
    $h$ & 150\,m & UAV flight altitude \\
    $\alpha$ & $90^{\circ}$ & Camera angle of view \\
    \midrule
    \multicolumn{3}{c}{\textbf{Power Model}} \\
    \hline
    $A$ & 1130.97 & Lift-induced power coefficient \\
    $B$ & 0.01353 & Parasitic power coefficient \\
    \midrule
    \multicolumn{3}{c}{\textbf{Time and Scheduling}} \\
    \hline
    $T_b$ & 5\,s & Time per \bez curve \\
    $T_f$ & 1\,s & Time between camera frames \\
    \bottomrule[1.5pt]
    \end{tabular*}
    \caption{Physical model parameters for fixed-wing UAV flight, camera configuration, and power consumption.}
    \label{tab:phys_params}
\end{table}

The scenarios for training and evaluation are procedurally generated, with parameters presented in Tab.~\ref{tab:phys_params}. Maps with a maximum side length of 2\,km are generated including 20 randomly distributed NFZs with a maximum side length of 400\,m. After the NFZs, 10 TZs with a maximum side length of 800\,m are generated, merged, and the NFZ areas are removed from the targets. After the map is generated, it is cropped to an area determined linearly by the difficulty parameter, with a randomized aspect ratio of a maximum of 3. 

The UAV parameters are similar to the ones from a Great Planes Avistar Elite fixed-wing trainer-type radio control aircraft. We assume a constant flight speed of $20$\,m/s, a flight altitude of 150\,m, and a maximum roll angle of 45\textdegree. With a camera angle of view of 90\textdegree, the UAV has a FoV side length of 300\,m. Through the power parameters $A= 1130.97$ and $B=0.01353$ from~\cite{theile2018uavee} the UAV requires 165\,W of power in level flight and 221\,W at the maximum roll angle. Every $T_b=5$\,s, a new \bez curve is generated, and every $T_f=1$\,s, a camera frame is captured.

The agents are trained using the curriculum on procedurally generated scenarios. First, the feasibility policy is trained using a parallelized version of the feasibility model for 1$\scriptstyle{\times 10^6}$ batches, requiring $\sim 9$\,h on an NVIDIA GeForce RTX 4090. Afterward, the objective policy is trained for 50$\scriptstyle{\times 10^6}$ interaction steps, requiring $\sim 60$\,h. Three agents were trained, and their performance is shown below.

\subsection{Results}

\subsubsection{Curriculum-based Training}

\begin{figure}
\centering
    \vspace{5pt}
    \begin{subfigure}{0.5\linewidth}
        \centering
        \includegraphics[width=0.9\textwidth]{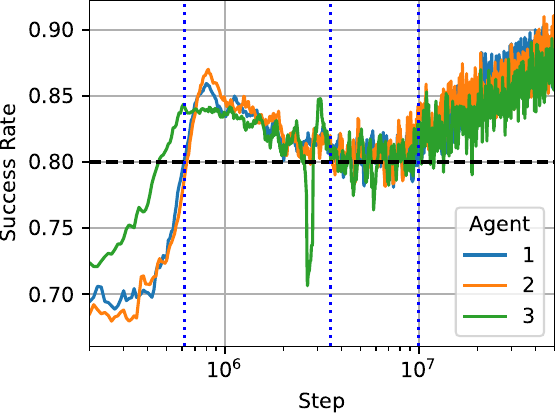}
        \caption{Success}
    \end{subfigure}%
    \begin{subfigure}{0.5\linewidth}
        \centering
        \includegraphics[width=0.9\textwidth]{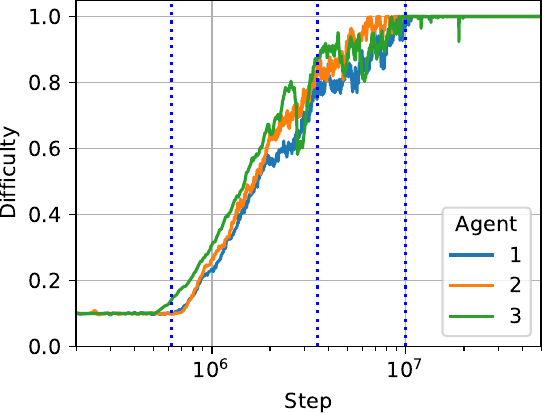}
        \caption{Difficulty}
    \end{subfigure}
    \begin{subfigure}{0.5\linewidth}
        \centering
        \includegraphics[width=0.9\textwidth]{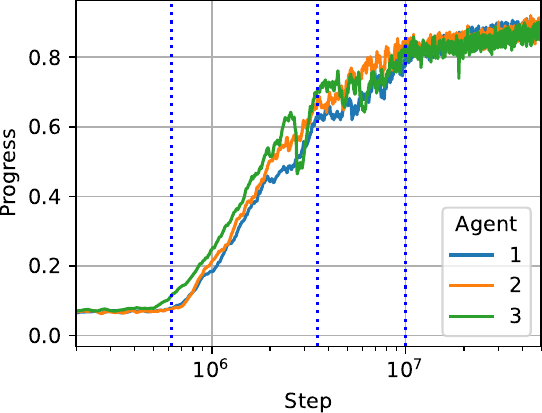}
        \caption{Progress}
    \end{subfigure}%
    \caption{Training curve of one agent using the curriculum, showing the increase of episodic difficulty dependent on the progress, using a log scale for training steps.}
    \label{fig:curriculum}
\end{figure}

The training progress of the three agents is shown in Fig.~\ref{fig:curriculum}, illustrating the effect of the curriculum. Four phases are visible, and the transitions are highlighted for agents 1 and 2: (i) From the start to around 6.0$\scriptstyle{\times 10^5}$ the agents exhibit success rates below the threshold $s_\text{th}=0.8$ and thus the difficulty does not increase above the minimum difficulty of $d_\text{min}=0.1$. (ii) The agents achieve increasing success rates, making the difficulty climb rapidly. (iii) From around 3.5$\scriptstyle{\times 10^6}$ the agents' success rate hovers only slightly above $s_\text{th}$ making the difficulty grow only slowly. (iv) Finally, at around 1.0$\scriptstyle{\times 10^7}$ steps the progress exceeds $s_\text{th}$ and the difficulty is set to 1 and the success rate slowly increases beyond the threshold value. 

The 3rd agent follows the same pattern but exhibits some instability. Specifically, at approx. 2.5$\scriptstyle{\times 10^6}$ steps the performs drops rapidly, to which the curriculum adapts by lowering the difficulty. Additionally, after 1.0$\scriptstyle{\times 10^7}$ steps, the 3rd agent occasionally drops in success rate, leading to dips in the difficulty. Overall, it can be observed that the curriculum adaptively adjusts the difficulty based on the agents' variable performances.

\subsubsection{Difficulty-dependent Performance}

\begin{figure}
    \centering
    \includegraphics[width=0.5\linewidth]{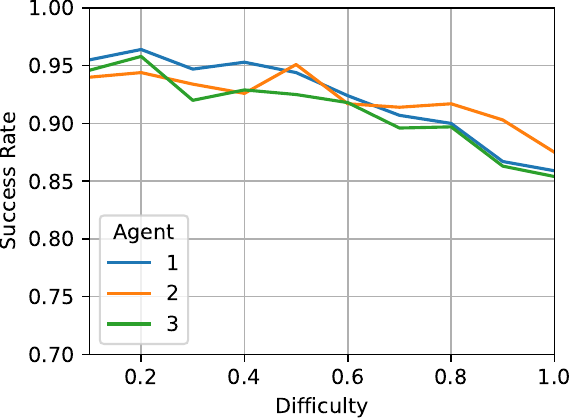}
    \caption{Success rates over difficulties for all agents.}
    \label{fig:results_over_difficulties}
\end{figure}

After training, the agents are evaluated on randomly generated scenarios of various difficulty levels. The results are shown in Fig.~\ref{fig:results_over_difficulties}. It can be seen that all agents perform similarly, with higher success rates at lower difficulty and a gradual decline in performance with increased difficulty. 

\subsubsection{Trajectories}

\begin{figure}[t]
  \centering
    \vspace{5pt}
    \begin{tabular}{cc}
      \includegraphics[width=0.45\linewidth]{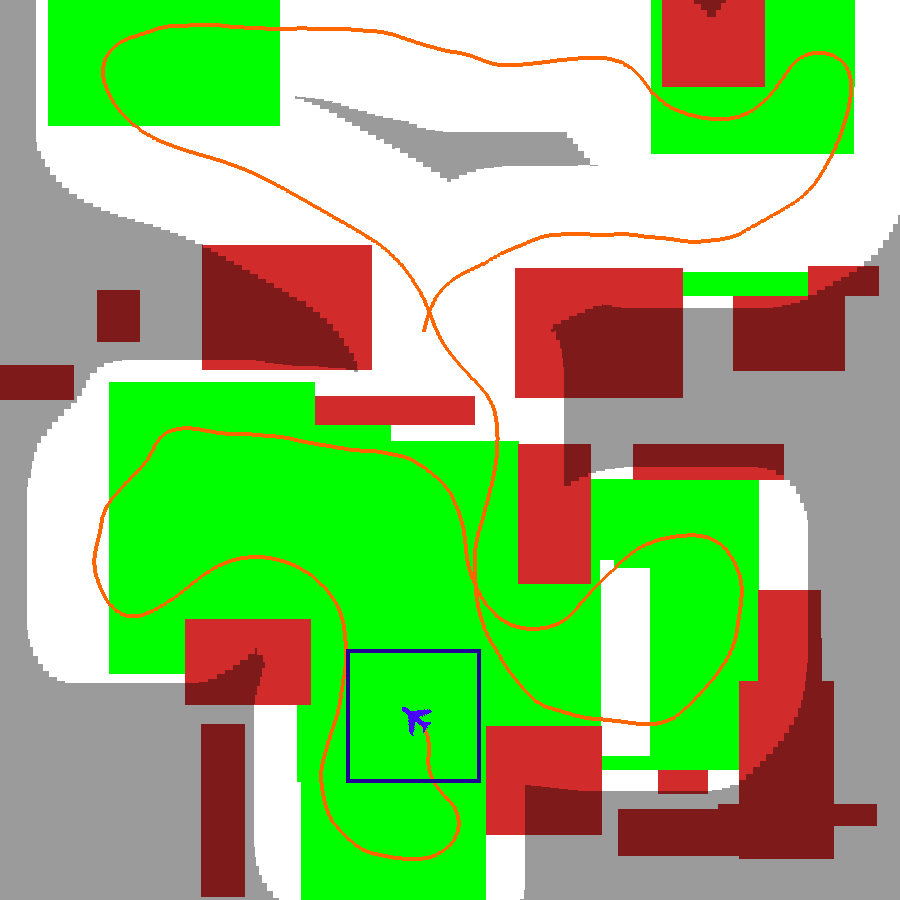} & 
      \includegraphics[width=0.45\linewidth]{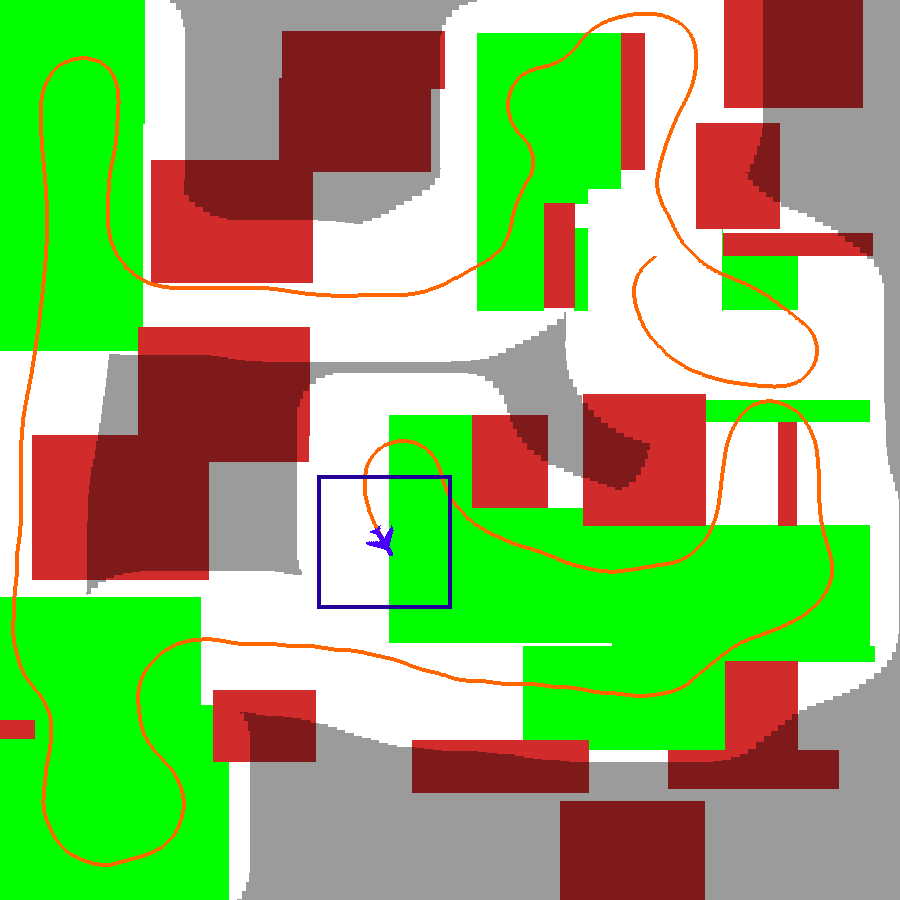} \\
      \includegraphics[width=0.45\linewidth]{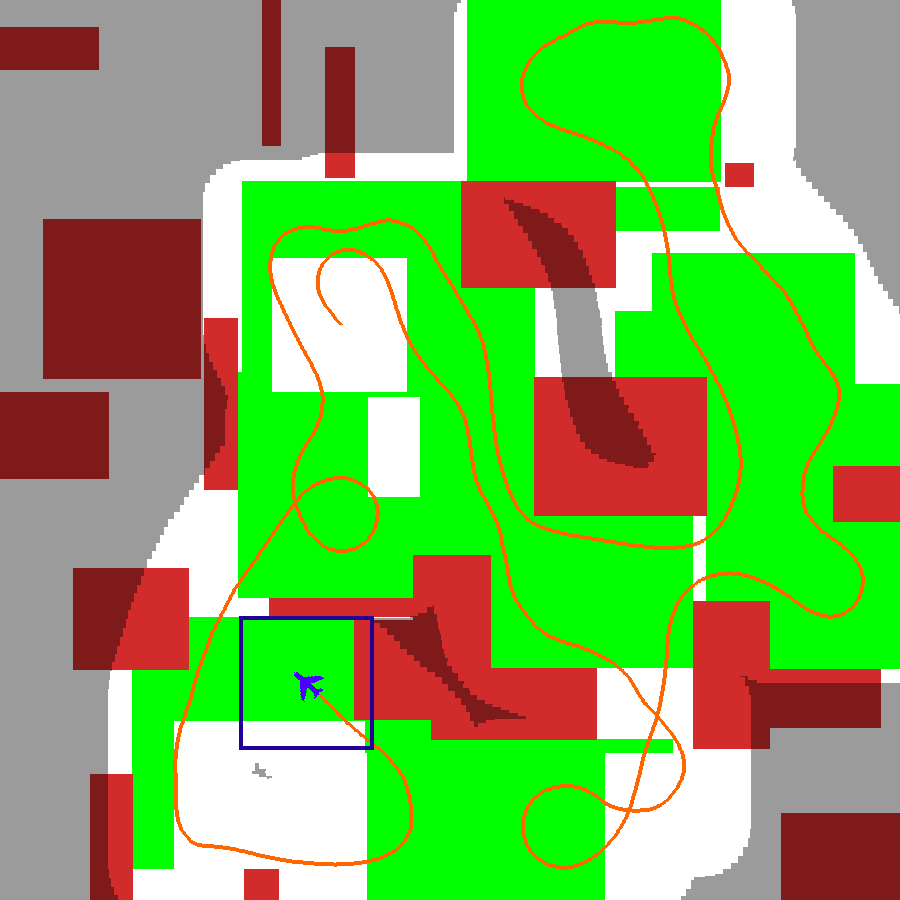} & 
      \includegraphics[width=0.45\linewidth]{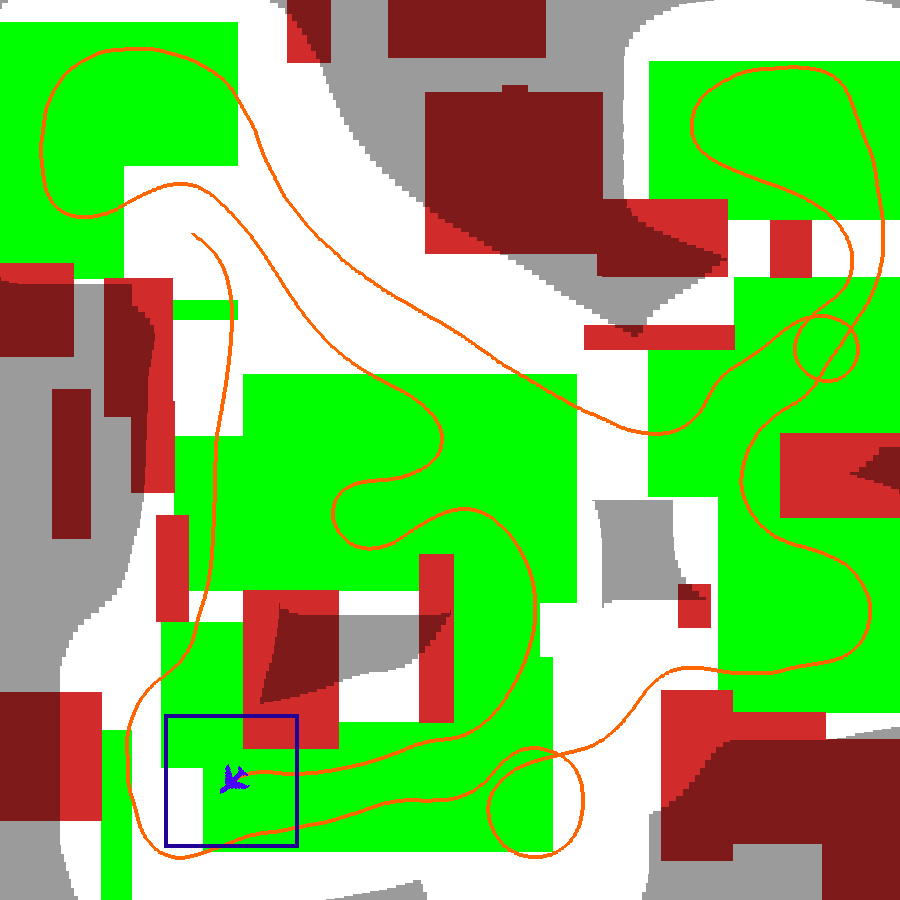} \\
    \end{tabular}
  \caption{Trajectory examples on procedurally generated maps.}
  \label{fig:trajectories_id}
\end{figure}

Fig.~\ref{fig:trajectories_id} shows selected trajectory plots of an agent solving procedurally generated scenarios. While most trajectories show smooth paths with little coverage overlap, some loops and trajectory intersections appear. These loops can usually be observed when the agent incorrectly estimated that it could cover TZs in a corner, but after realizing that the estimation was incorrect, it has to double back. Besides occasional inefficiencies, the agent overall shows well-planned trajectories that cover the entire target on the randomly generated scenarios it has never seen before.

To show out-of-distribution generalization, we also tested the agent on hand-crafted maps similar to Manhattan32 and TUM50 in~\cite{theile2023learning}. In these maps, NFZs are manually designed, while TZs remain randomly generated. It can be seen that the agent solves these out-of-distribution maps very smoothly, indicating that the procedurally generated scenarios are challenging enough to allow the agent to generalize to more realistic hand-crafted maps.  

\begin{figure}[t]
  \centering
    \begin{tabular}{cc}
      \includegraphics[width=0.45\linewidth]{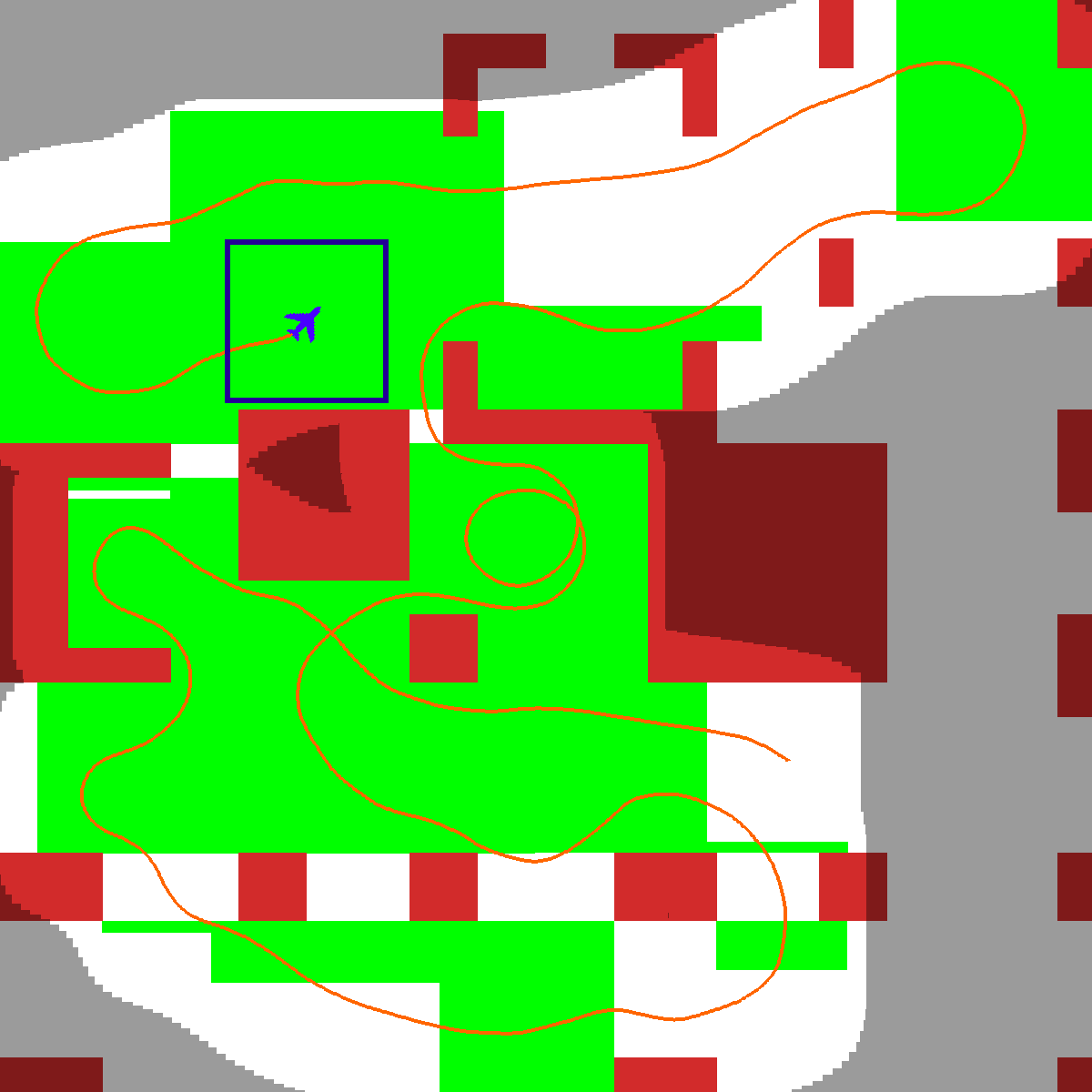} & 
      \includegraphics[width=0.45\linewidth]{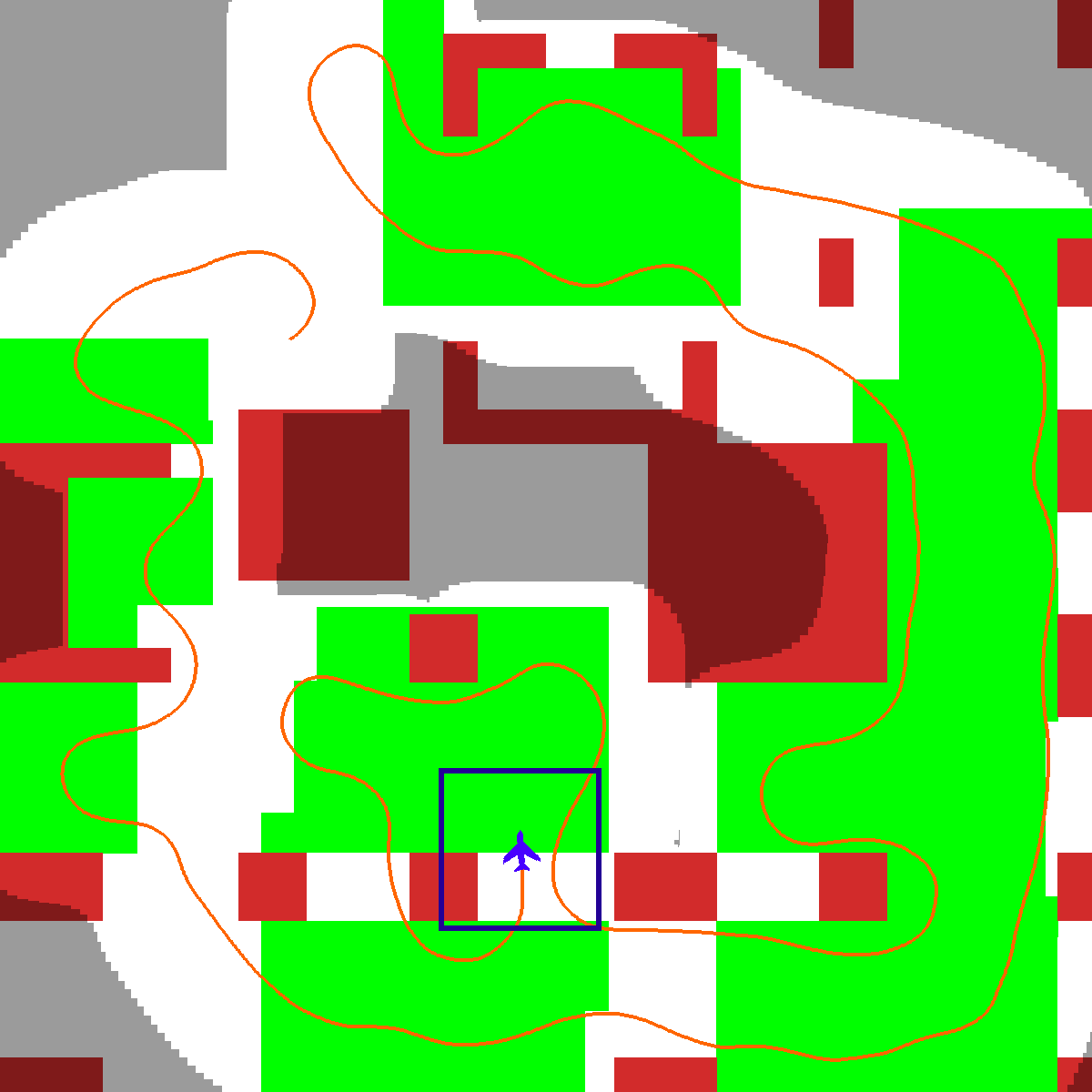} \\
      \includegraphics[width=0.45\linewidth]{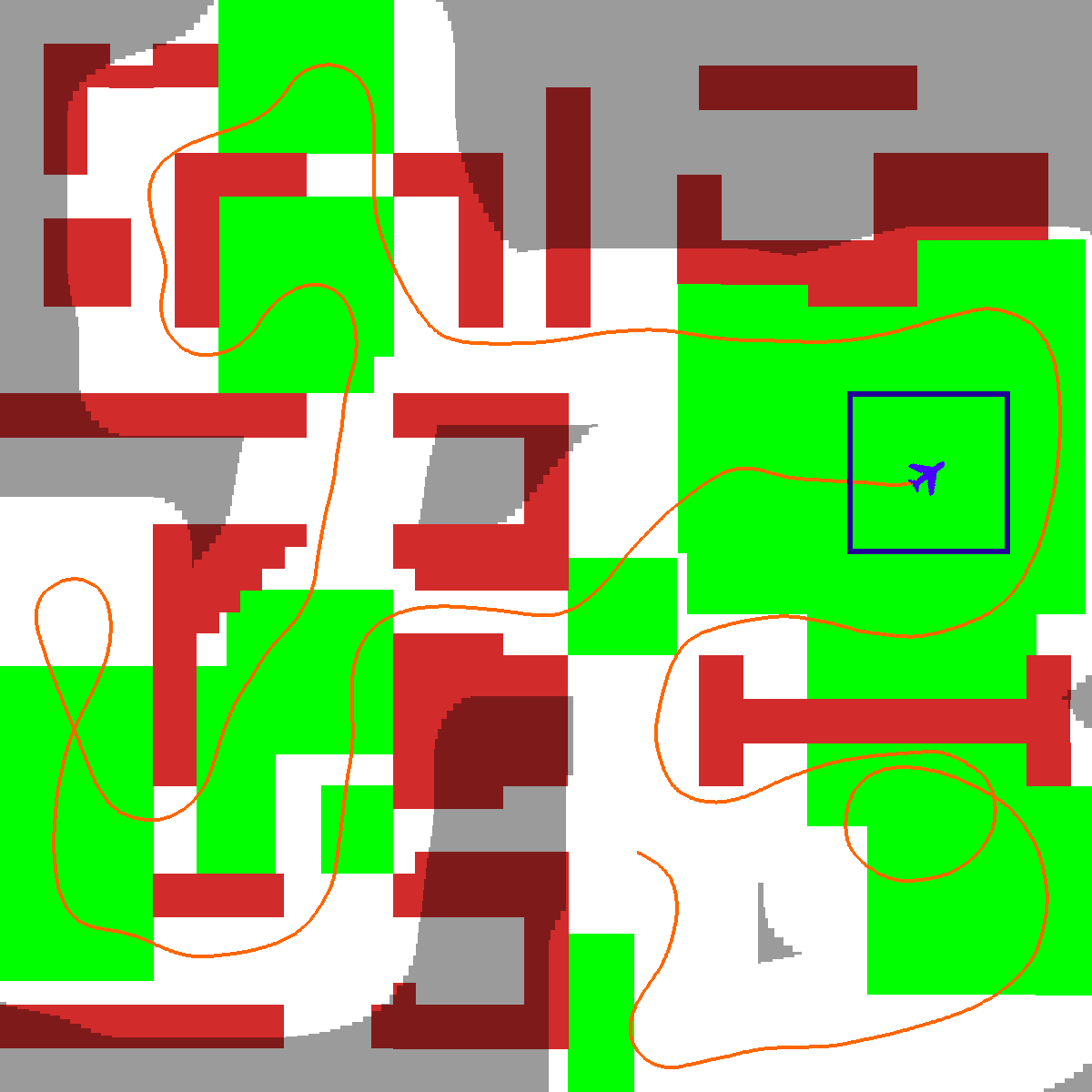} & 
      \includegraphics[width=0.45\linewidth]{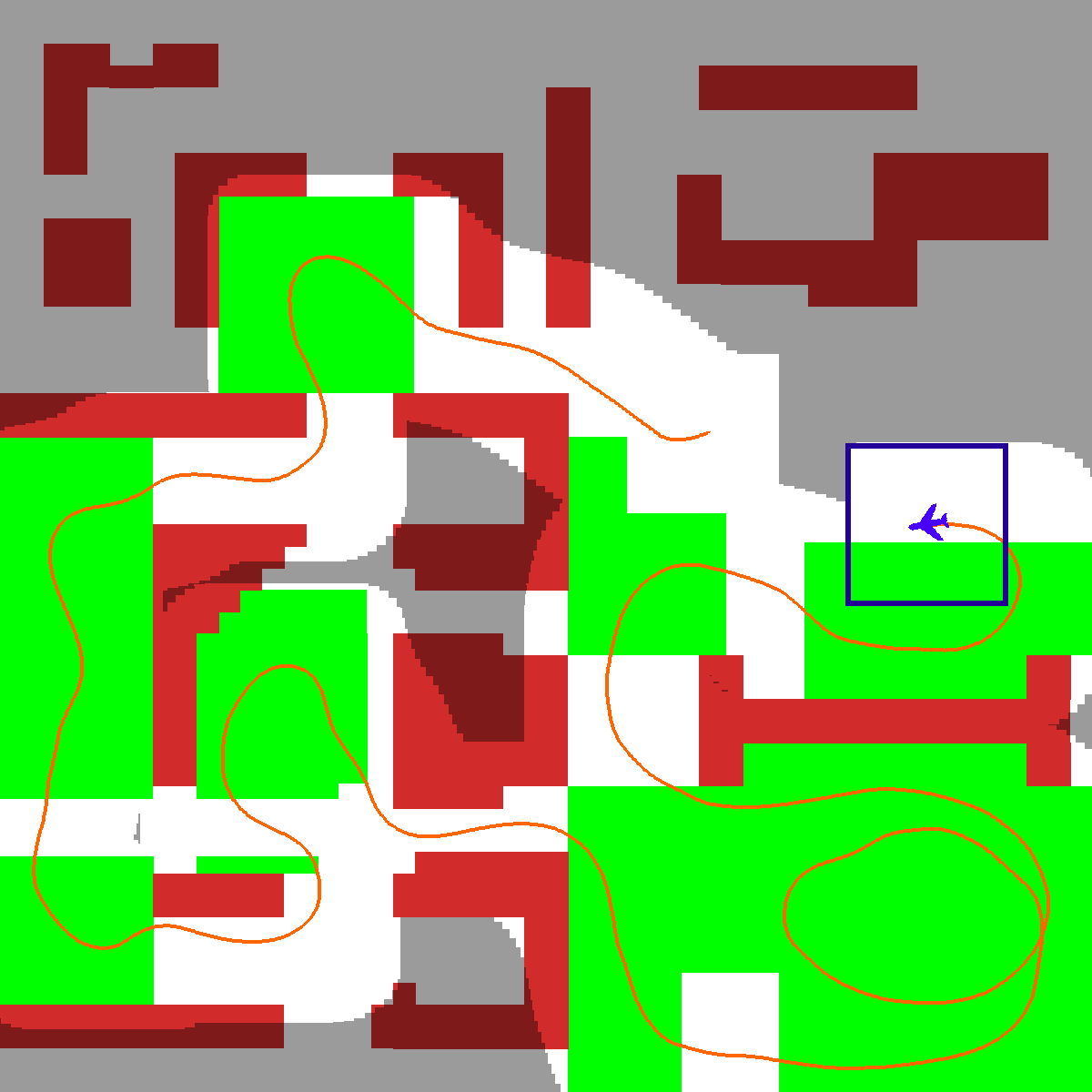} \\
    \end{tabular}
  \caption{Trajectory examples on two hand-crafted maps.}
  \label{fig:trajectories_ood}
\end{figure}

\section{Conclusion and Future Work}

This paper presented a novel approach for continuous-world UAV CPP using DRL with a self-adaptive curriculum. Using a set-based representation of the environment and an attention-based DNN, the agent can learn complex spatiotemporal relationships and make long-term decisions. By formulating the trajectory as consecutive \bez curves and introducing a corresponding action space, the DRL agent can learn smooth trajectories that are feasible for an autopilot. The agent can learn to solve large environments through the self-adaptive curriculum without any reward shaping. Experiments showed that the agent can solve scenarios generated from the same distribution and solve out-of-distribution hand-crafted maps. Overall, the formulation and methodology bring the application of DRL closer to real-world UAV CPP.

Despite the advances, multiple problems need to be addressed. The current environment description only allows for axis-aligned rectangles, which is a limiting factor. As the next step, we will investigate how to describe and observe the environment through arbitrary polygons. Additionally, since the feasibility policy is trained, it sometimes exhibits irregularities that hamper the learning of the objective policy. Preliminary experiments showed that exploiting symmetries of the environment~\cite{theile2024equivariant} can mitigate this problem.

\bibliographystyle{ieeetr}
\bibliography{bib}

\end{document}